\documentclass[10pt,letterpaper,twocolumn]{article}
\pdfoutput=1
\usepackage[latin9]{inputenc}
\usepackage{color}
\usepackage{float}
\usepackage{amsmath}
\usepackage{amssymb}

\usepackage{graphicx}

\makeatletter

\pdfpageheight\paperheight
\pdfpagewidth\paperwidth

\providecommand{\tabularnewline}{\\}
\floatstyle{ruled}
\newfloat{algorithm}{tbp}{loa}
\providecommand{\algorithmname}{Algorithm}
\floatname{algorithm}{\protect\algorithmname}

\usepackage{cvpr}\usepackage{times}\usepackage{epsfig}

\cvprfinalcopy 

\def\cvprPaperID{1134} 

\ifcvprfinal\fi

\makeatother

\begin{document}

\title{Face-space Action Recognition by Face-Object Interactions}

\author{Amir Rosenfeld and Shimon Ullman\\  Weizmann Institute of Science   \\  Rehovot, 7610001, Israel \\ {\tt \{amir.rosenfeld,shimon.ullman\}@weizmann.ac.il}}
\maketitle
\begin{abstract}
Action recognition in still images has seen major improvement in recent
years due to advances in human pose estimation, object recognition
and stronger feature representations. However, there are still many
cases in which performance remains far from that of humans. In this
paper, we approach the problem by learning explicitly, and then integrating
three components of transitive actions: (1) the human body part relevant
to the action (2) the object being acted upon and (3) the specific
form of interaction between the person and the object. The process
uses class-specific features and relations not used in the past for
action recognition and which use inherently two cycles in the process
unlike most standard approaches. We focus on face-related actions
(FRA), a subset of actions that includes several currently challenging
categories. We present an average relative improvement of 52\% over
state-of-the art. We also make a new benchmark publicly available.
\end{abstract}

\section{Introduction}

\begin{figure}
\includegraphics[width=1\columnwidth]{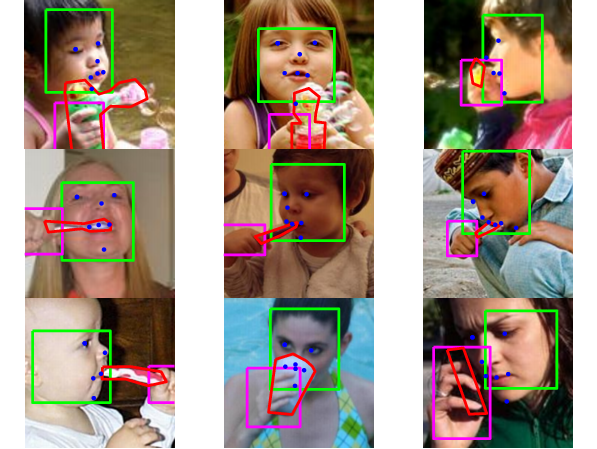}

\protect\caption{\label{fig:fra-db}Sample annotations from FRA-db, showing \textcolor{magenta}{hands},
\textcolor{green}{faces}, \textcolor{blue}{facial-landmarks} and \textcolor{red}{action
objects}. Best viewed online.}
\end{figure}

Recognizing actions in still images has been an active field of research
in recent years, with multiple benchmarks appearing \cite{pascal-voc-2012,delaitre2010recognizing,sadeghi2011recognition,ICCV11_0744}.
It is a challenging problem since it combines different aspects including
object recognition, human pose estimation, and human-object interaction.
Action recognition schemes can be divided into transitive vs. intransitive,
and to methods using dynamic input vs. recognition from still images.
Our focus is on the recognition of transitive actions from still images.
Such actions are typically based on an interaction between a body
part and an action-object (the object involved in the action). 
\begin{figure}
\includegraphics[width=1\columnwidth]{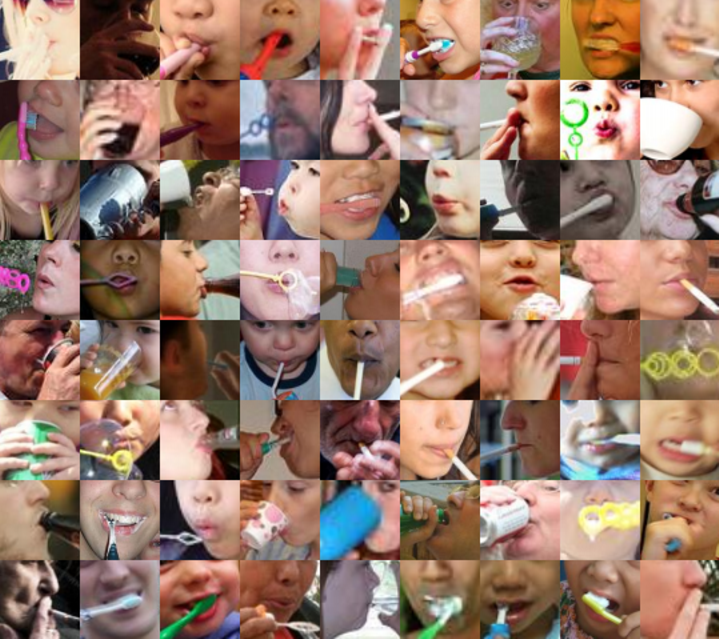}

\protect\caption{\label{fig:action-zoomed-in}Inferring a transitive action in an image
is often made possible by using shape properties and relations that
are highly informative for recognition, yet small and specific to
a particular action. All actions involve interactions between the
mouth region and action-object, the different actions often differ
in small local detail of the object or its interaction with the mouth.}
\end{figure}
 In a cognitive study of 101 frequent actions and 15 body parts, hands
and faces (in particular mouth and eyes) were by far the most frequent
body parts involved in actions \cite{maouenea2008body}. It is interesting
to note that in the human brain, actions related to hands and faces
appear to play a special role: neurons have been identified in both
physiological and fMRI studies \cite{brozzoli2009peripersonal,ladavas1998visual},
which appear to represent 'hand space' and 'face space', the regions
of space surrounding the hands and face respectively. Here, we focus
on actions in face-space, in particular involving the mouth as a body
part and different actions and objects, including drinking, smoking,
brushing teeth and blowing bubbles. We refer to them as face-related
actions (FRA). In cases where action-objects can be easily detected,
the task may be guided by object detection. In some other classes
which take place in specific settings (fishing, cutting trees), the
task may be solved by recognizing the context or scene rather than
the action itself. However, for many actions, the action-object can
be difficult to recognize, and the surrounding context will not be
directly informative. Such actions include (among many others) the
mentioned face-related actions. As illustrated below, to recognize
these actions it is often required to extract and use in the classifier
detailed features of the interaction between a person and an action-object
and analyze fine details of the action object being manipulated. This
includes both detecting the location of the action object and applying
a fine analysis of this object with respect to the person as well
as the appearance of the object itself. Common to these action classes
are several difficulties: The action objects may be quite small (relative
to objects in action classes in which classifiers achieve higher performance),
they may appear in various forms or configurations (e.g, cups, straws,
bottles, mugs for drinking, different orientations for toothbrushes)
and are often highly occluded, to the degree that they occupy only
a few pixels in the image. Moreover, often a very small region in
the image is sufficient to discern the action type - even if it contains
only a part of the relevant object, as in Fig. \ref{fig:action-zoomed-in}.
Note how a few pixels are required to tell the presence of an action
object, its function and interaction with the face. Subtle appearance
details and the way an object is docked in the mouth serve to discriminate
between \eg smoking (first row, first image) and brushing teeth (2nd
row, 3rd image). The approach described here includes the following
contributions:
\begin{itemize}
\item A new and robust facial landmark localization method
\item Detection and use of a\textit{ interaction} features, a representation
of the relation between action-objects and faces
\item Accurate localization of action objects, allowing the extraction of
informative local features.
\item Large improvement in state-of-the-art results for several challenging
action classes
\end{itemize}
We augment the Stanford-40 Actions dataset \cite{ICCV11_0744} by
adding annotations for classes of drinking, smoking, blowing bubbles
and brushing teeth, which include the locations of several facial
landmarks, exact delineation of the action objects in each image and
of the hand(s) involved in the action. We dub this dataset FRA-db,
for Face Related Actions. We will make this dataset available as a
new challenge for recognizing transitive actions. See Fig. \ref{fig:fra-db}
for some sample annotations. 

The rest of the paper is organized as follows: Section \ref{sub:Related-Work}
is dedicated to related work. In Section \ref{sec:Approach} we explain
in detail our own method. Section \ref{sub:Adding-to-the} contains
some more details about the newly collected annotations in FRA-db.
Finally, in Section. \ref{sec:Experiments} we show experimental results,
illustrating the effectiveness of our approach. 


\subsection{Related Work\label{sub:Related-Work}}

Recent work on action recognition in still images can be split into
several approaches. One group of methods attempt to produce accurate
human pose estimation and then utilize it to extract features in relevant
locations, such as near the heads or hands, or below the torso \cite{yao2012action,desai2012detecting}. 

Others attempt to find relevant image regions in a semi-supervised
manner: \cite{eth_biwi_00828} find candidate regions for action-objects
and optimize a cost function which seeks agreement between the appearance
of the action objects for each class as well as their location relative
to the person. In \cite{sener2012recognizing} the objectness \cite{alexe2010object}
measure is applied to detect many candidate regions in each image,
after which multiple-instance-learning is utilized to give more weight
to the informative ones. Their method does not explicitly find regions
containing action objects, but any region which is informative with
respect to the target action. In \cite{yao2011combining} a random
forest is trained by choosing the most discriminative rectangular
image region (from a large set of randomly generated candidates) at
each split, where the images are aligned so the face is in a known
location. This has the advantage of spatially interpretable results. 

Some methods seek the action objects more explicitly: \cite{ICCV11_0744}
apply object detectors from Object-Bank \cite{li2010object} and use
their output among other cues to classify the images. Recently, \cite{DBLP:conf/icmcs/LiangWHL14}
combine outputs of stronger object detectors together with a pose
estimation of the upper body in a neural net-setting and show improved
results, where the object detectors are the main cause for the improvement
in performance. 

In contrast to \cite{eth_biwi_00828}, we seek the location of a specific,
relevant body part only. For detecting objects, we also use a supervised
approach, but unlike $\eg$ \cite{DBLP:conf/icmcs/LiangWHL14}, we
represent the location of the action-object using a shape mask, to
enable the extraction of rich \textit{interaction} features between
the person and the object. Furthermore, we use the fine pose of the
human (i.e, facial landmarks) to predict where each action object
can, or cannot, be, in contrast to \cite{delaitre2011learning}, who
use only relative location features between body parts and objects.
We further explore features specific to the region of interaction
and its form, which are arguably the most critical ones to consider.

\section{Approach\label{sec:Approach}}

The overall goal is to classify an image $I$ into one of $C$ action
classes. In this section we describe briefly the main stages in the
process, with more details in subsequent sections. We use for the
classification features that come from three components:
\begin{enumerate}
\item The body part relevant to the action being performed
\item The action object
\item The specific form of interaction between the body part and action
object
\end{enumerate}
For each of these components, we extract and learn discriminative
features to discern between the different action classes. To this
end, we must first detect accurately both the body part and the action-object,
which will enable us to examine their interaction. Since we deal with
face-related actions, we first detect faces and extract facial landmarks,
as described in Section \ref{sub:Facial-Landmarks}. In the following,
let \textbf{$F=(X,S,L)$ }be a face detected in the image \textbf{$I$},
where \textbf{$X$} are the image coordinates of the rectangle output
by the face detector, $S$ is the face detection score, and $L=(x_{k},y_{k})_{k=1}^{n}$
are $n$ coordinates of detected facial landmarks. We crop out a rectangular
region centered as $X$ but with larger dimensions (1.5 throughout
our experiments) and resize it to a constant-sized image, denoted
by $I_{F}$. In addition we crop a square sub-image $I_{m}$ centered
on the mouth (one of the detected facial landmarks, see below) proportional
to the size of the original face (1/3 of the height and width, found
adequate to include both corners of the mouth and its upper and lower
lips). To detect the action object, we generate a pool $R$ of candidate
regions inside $I_{F}$, in a predicted location (using ANN-star,
below). The regions $r\in R$ serve as candidates for action-objects.
We then evaluate for the face \textbf{$F$} the following score function
for each class $c\in1,..C$:

\begin{equation}
S_{c}(I,F)=\eta_{c}^{F}(I_{F})+\eta_{c}^{M}(I_{M})+\mbox{\ensuremath{\gamma}}\max_{r\in\mathcal{R}}(\mathcal{\mathcal{\eta}}_{c}^{Int}(r,F)+\eta_{c}^{Obj}(r))\label{eq:energy function}
\end{equation}

Equation \ref{eq:energy function} contains 4 components; we next
describe them briefly and point to the sections elaborating on the
features used for each component. 
\begin{itemize}
\item $\eta_{c}^{F}(I_{F})$ and $\eta_{c}^{M}(I_{M})$ are scores based
on appearance features extracted from the face and mouth sub-images
$I_{F}$ and $I_{M}$ respectively (Section. \ref{sub:Face-and-Mouth})
\item $\eta_{c}^{Int}(r,F)$ is a score function based on \textit{interaction
}features, that is, the compatibility of the geometric relations of
a region $r$ w.r.t to $F$ for the expected action class (Section
\ref{sub:Person-Object-Interaction}). This requires accurate detection
of facial landmarks, which is described in Section \ref{sub:Facial-Landmarks}.
\item $\eta_{c}^{Obj}(r)$ expresses a score based on appearance and shape
features extracted from the candidate action-object region $r$ in
the image $I_{F}$ ( Section \ref{sub:Detecting-Action-Objects})
\end{itemize}
To evaluate Eq. \ref{eq:energy function} we compute $\eta_{c}^{F}(I_{F})$
and $\eta_{c}^{M}(I_{M})$ for each face and $\mathcal{\mathcal{\eta}}_{c}^{Int}(r,F)+\eta_{c}^{Obj}(r)$
for each candidate region $r$. We pick the candidate object region
$r$ that produces the maximal value to obtain the final score $S_{c}(I,F)$.
All terms $(\cdot)_{c}$ are learned in a class-specific manner from
annotated images. At training time, we use manually annotated locations
of the faces, their facial landmarks, as well as the exact location
of the action-object, given as a binary mask. The next sections elaborate
on how we detect faces and their landmarks, generate candidate regions
and extract the relevant features.

It is worth noting that in our approach, the types of features, and
the image regions from which they are extracted, are progressively
refined at successive stages. We first constrain the region of the
image to be analyzed for action-object by finding the relevant body-part
(face), then a sub-part (the mouth) that further refines the relevant
locations, to extract features (such as region-intersections and bounding
ellipses), which are not extracted at other image locations. The training
procedure is described in Section \ref{sub:Training}.

\subsection{Face and Mouth Features\label{sub:Face-and-Mouth}}

Assume we are given the bounding box of a face and the center of the
mouth. Both are either extracted automatically in testing (Sec. \ref{sub:Facial-Landmarks})
or given manually in training. We extract a feature representation
of the entire face area $I_{F}$ by using the fc6 layer of a convolutional
neural net, as in \cite{DBLP:dblp_journals/corr/RazavianASC14}, which
produced the best results in preliminary comparisons, but we use the
network defined in\textbf{ imagenet-vgg-m-2048} from \cite{Chatfield14},
since it has a good trade of between run-time, output dimensionality
and performance. Denote these features as fc6 features; they are produced
for an image ($\eg,$$I_{F})$ fed into the neural net after the proper
resizing and preprocessing. This produces the features used to train
the score function $\eta_{c}^{F}(I_{F})$. Similarly, We extract features
from $I_{M}$, a square region around the mouth area, producing features
to train $\eta_{c}^{M}(I_{M})$.

\subsection{\label{sub:Detecting-Action-Objects}Action Object Features}

We use segmentation to produce action-object candidates, augmented
with a method for cases where the object is not included in the result
of the segmentation. We do this around a predicted location, guided
by an ANN-Star model (Section \ref{sub:ANN-Star}). We first produce
a large number of candidate regions, then discard regions based on
some simple criteria and finally extract a set of features from each
region, as is described next.

We produce a rich over-segmentation of the image by applying a the
recent segmentation of \cite{arbelaez2014multiscale}. It is both
relatively efficient and has a very good F-measure. We use the segments
produced (with some extensions, see below) by this method as candidates
for the action-object. Before applying the segmentation, we crop the
image around the detected face as in Section \ref{sec:Approach},
to produce $I_{F}.$ The object may extend beyond this, but we are
interested in the part interacting with the face, as the region of
interaction bears the most informative features. The number of regions
produced in this way varies between a few hundred up to about 5,000.

Next, we discard regions based on the following criteria: the area
is (a) too small (30 pixels; as there were no ground-truth regions
below this size in the training set), or (b) too large relative to
the area of $I_{F}$, \ie, more than 50\%, determined from the training
set.

For appearance, we use the same fc6 features as described in Section
\ref{sub:Face-and-Mouth}, extracted from the rectangular image window
containing the region.

To extract shape features, we first compute a binary mask for each
candidate region, in the coordinate frame of $I_{F}$. We create our
pool of features in the following manner:
\begin{enumerate}
\item The binary mask is resized to a 7x7 image, producing an occupancy
mask in a coordinate system centered on the face: this is to capture
the distribution of locations of the action object (per class) relative
to the face.
\item The mask is cropped using its minimal bounding rectangle, resized
to 64x64 pixels (using bilinear interpolation) and a 8x8 HOG descriptor
is extracted, as a representation of the shape of the region.
\item The following shape properties: the major and minor axis of the ellipse
approximating the shape of the region, its total area, eccentricity,
and orientation. 
\end{enumerate}
We concatenate all of these features along with the appearance features
and denote them by $V_{Obj}(r).$

\subsubsection{Finding objects overlooked by segmentation}

The pool of candidate regions produced by the segmentation is useful
in most cases, but it will not always include the action-object as
one of the proposed segments. The approach we propose to deal with
such cases is to search in addition for regions defined by local features
that appear with high probability in the action object, such as parallel
contours (for straws and cigarettes), and elliptical contours (for
cup-rims) and others. These are relatively complex features, but the
localization of the mouth as a likely object location confines the
search to a limited region. We used in particular the presence of
parallel line segments \cite{KovesiMATLABCode}. The quadrangles formed
by the 4 corners of these pairs of line segments are kept in the same
pool of regions created by the segmentation, and are used in both
training and testing. As can be seen in Fig. \ref{fig:Interp-image}
(top row, second from left), these additional regions aid in capturing
informative regions which are missed or given a low score by the original
segmentation.\textbf{ }The additional regions produced by this method
are added to the pool of regions $R$ and treated exactly as other
regions for the remainder of the process.

\subsection{\label{sub:Person-Object-Interaction}Face-Object Interaction}

We describe next how we model person-object interaction. As described
above, the exact relative position and orientation of the action-object
with respect to the face part (mouth) can be crucial for classifying
actions. Our method therefore learns informative features of the person-object
interactions and uses them during classification. The interaction
features of each candidate region $r$ with respect to the face $F$
are computed using the following measurements:
\begin{enumerate}
\item The expected location of the object center: this is computed by the
output of an ANN-Star model (Section\ref{sub:ANN-Star}) trained to
point to this location, given features extracted from $I_{F}$. The
ANN-Star model produces a probability map over the image $I_{F}$,
denoted by $P_{c}^{l}(x,y)$. 
\item A saliency map by \cite{zhu2014saliency}, to create $P^{s}(x,y)$,
which is proved helpful for avoiding background objects. 
\item The maximal and minimal distances of any point on the region w.r.t
each of the 7 detected facial landmarks, for a total of 14 distances.
\item The relations of the action-object to the facial landmarks are often
informative for recognition. For each region $r$ and facial landmark
$p$, we create a log-polar binning centered on $p$, and count the
of pixels of to $r$ in each bin. This encodes not only the rough
minimal distance (as in (3)) from $r$ to $p$ but also how much of
the surrounding area of $p$ is covered by $r$. 
\item We extract three measures of the overlap between $r$ and the original,
(not inflated) bounding box of $F$. Let $R_{F}$ be the rectangular
region of the face $F$. we measure the relative intersection areas
$\frac{|r\cap R_{F}|}{|R_{F}|},$$\frac{|r\cap R_{F}|}{|r|}$ as well
as the overlap score $\frac{|r\cap R_{F}|}{|r\cup R_{F}|}.$ The relative
area of each region in the other ($r$ in F and $F$ in $r)$ are
useful properties to determine if there is any interaction between
$F$ and the region $r$.
\end{enumerate}
The interaction features are now defined as follows: The average of
$P_{c}^{l}(x,y)$ in $r$, the average value of $P^{s}(x,y)$ in $r$,
the 14 distances as in (3) above, the log-polar representations, per
landmark, in (4) and the 3 relative overlap properties as mentioned
in (5). The concatenation of these interaction features is denoted
by $V_{Int}(r)$.

\subsection{ANN-Star\label{sub:ANN-Star}}

Throughout the algorithm, we use in a uniform manner a common component,
the so-called ANN-Star model, akin to the method in \cite{leibe2004combined},
to predict the location of a target relative to a reference. This
is done for both facial landmark detection (Section \ref{sub:Facial-Landmarks})
and for action-object localization (Section \ref{sub:Person-Object-Interaction}).
We use the model as a non-parametric way of localizing specific objects
or points of interest within objects (\ie, facial landmarks), as
it performed well in our comparisons given relatively few training
examples with high variability. We briefly describe the model we used.
Let $(I_{j})_{j=1}^{N}$ be $N$ training images produced by cropping
out rectangular sub-images around the face and resized to a constant
size, as is in the previous section. Let $P_{j}=(x_{j},y_{j})$ be
the centers of objects of interest (a single object in each training
image) in $I_{j}$. We extract SIFT features densely at a single scale
from $I_{j}$, and sample randomly $K_{f}$ of them using weighted
sampling (without replacement). The weight for sampling a descriptor
centered on a point $u=(x,y)$ is computed as $w_{u}=G_{u}e^{-\|u_{1}-P_{j}\|_{2}/\sigma}$,
where $G_{u}$ is the image gradient at point $u$; this causes the
sampled features to be with high probability near $P_{j}$ and on
non-smooth patches. We sample $K_{f}=100$ patches from each training
image, a broad range of $K_{f}$ produces similar results. For each
sampled point in a training image $I_{j}$ we record $d_{i}^{j}=(f_{i}^{j},\Delta_{i}^{j}$)
where $f_{i}$ is the SIFT descriptor and $\Delta_{i}$ is the offset
of the patch center from the target center $P_{j}$. $f_{i}$ is normalized
using RootSIFT \cite{arandjelovic2012three}. We refer to all the
former descriptor as $\mathcal{X}=(f_{i}^{j})_{i=1..K_{f},j=1..N}$.
They are stored in a kd-tree \cite{muja_flann_2009} for fast retrieval.
For a test image $I_{t}$ (cropped around a face as in training) we
extract SIFT descriptors densely but without sampling, unlike the
training phase and normalize them using RootSIFT as well. For each
extracted feature $f_{t}$ we find the nearest-neighbor $f_{n}\in\mathcal{X}$
and cast a vote to the offset $\Delta_{n}$ recorded for $f_{n}$,
proportional to $e^{-\|f_{t}-f_{n}\|_{2}/\sigma_{2}}$. Since both
testing/training images are scaled with respect to the face, we can
vote in a single scale. The result is a pixel-wise heat-map (normalized
to sum = 1), to be used in subsequent steps of the method. $\sigma_{1}=10$
and $\sigma_{2}=.1$ were kept constant throughout the experiments.

\subsection{Facial Landmark Detection\label{sub:Facial-Landmarks}}

Images of actions involving the face naturally contain many occlusions
(as do other transitive actions, considering the respective body parts).
This is a challenging setting for facial landmark detection. To overcome
this difficulty, we have employed a method which is conceptually simple
but showed empirically superior results to publicly available methods,
including \cite{zhu2012face}. We extract $n=7$ facial landmarks,
which are: the left eye center, right eye center, left \& right corners
of mouth, mouth center, tip of nose and chin.

\begin{figure}
\begin{centering}
\includegraphics[width=1\columnwidth]{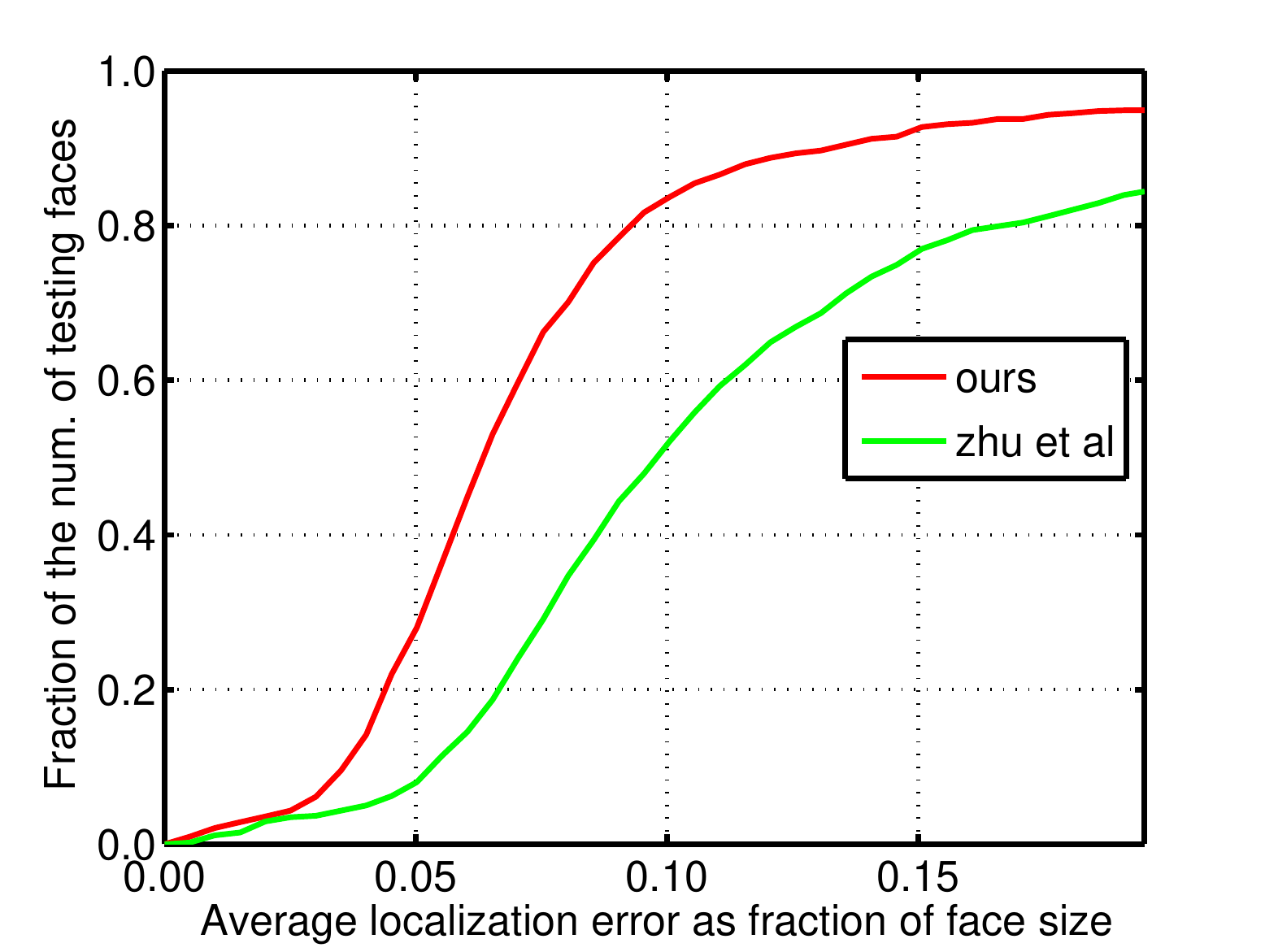}
\par\end{centering}

\begin{centering}
\includegraphics[width=0.4\columnwidth]{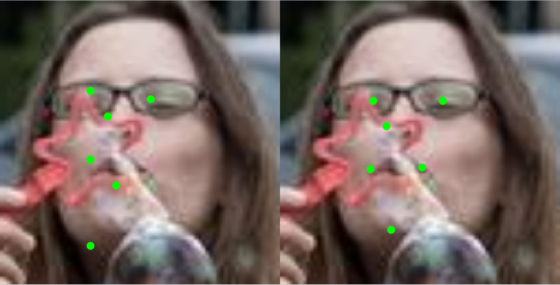}\includegraphics[width=0.4\columnwidth]{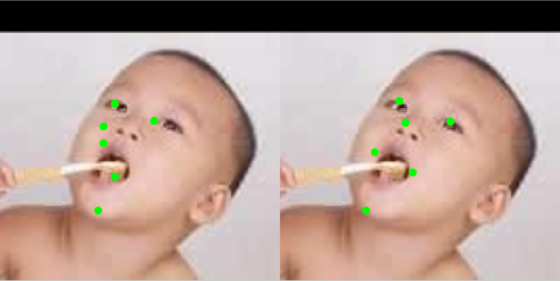}
\par\end{centering}

\begin{centering}
\includegraphics[width=0.4\columnwidth]{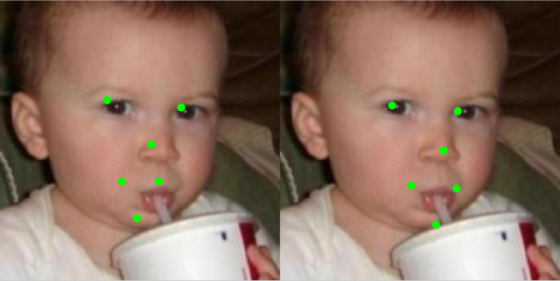}\includegraphics[width=0.4\columnwidth]{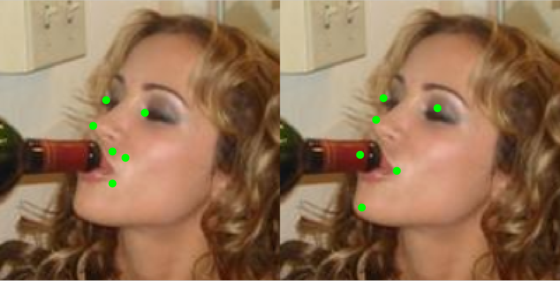}
\par\end{centering}

\begin{centering}
\includegraphics[width=0.4\columnwidth]{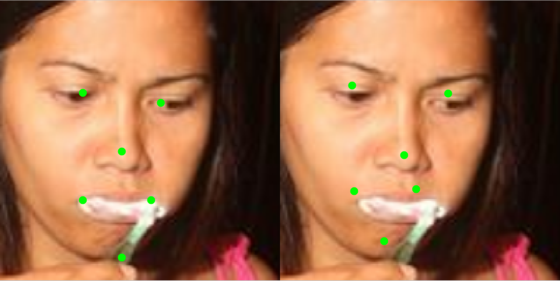}\includegraphics[width=0.4\columnwidth]{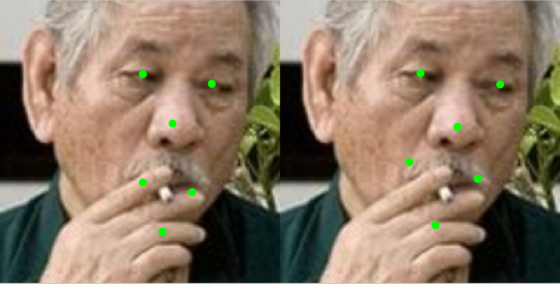}
\par\end{centering}

\protect\caption{\label{fig:landmark localization results}Top : comparison of our
landmark localization vs. Zhu et al \cite{zhu2012face}, over the
images in FRA-db. The plot shows the fraction of images for which
there is a mean error (normalized relative to face size) of up to
a given value. Bottom : some examples on challenging cases (in each
image pair, left is produced by Zhu et al and right is our results).
The bottom (last two) pair show examples where \cite{zhu2012face}
outperforms our scheme.}
\end{figure}

We produce for each facial landmark two hypotheses and then combine
them into a single result, as follows: given a large corpus of faces
with annotated facial landmarks \cite{tugraz:icg:lrs:koestinger11b},
we first apply a face detector \cite{mathias2014face} on the ground-truth
face images to align the faces in a consistent manner. We discard
faces whose score was too low (2.45) or did not contain more than
80\% of the landmarks in the ground-truth face. This leaves us with
a set of training faces $\mathcal{T}$ and their ground-truth landmarks.
Given a new test image I, we detect the face $F$. We then seek the
$K$ nearest neighbors of the $F$ in $\mathcal{T}$, where we use
the $L_{2}$ distance over HOG features extracted from both the $F$
and each image in $\mathcal{T}.$ This is done using a kd-tree for
efficiency. We then produce two sets of hypotheses for all landmarks:
\begin{itemize}
\item The first set is obtained by copying and transforming the location
of each required landmark from the $K$-nearest neighboring faces.
This produces a distribution of landmark-candidates for $F$, where
we use KDE (with $\sigma=20$) in order to find the maximum of the
distribution. We compute the distribution for each landmark independently.
Denote the locations found by this method as $(L_{i}^{g})_{i=1}^{n}$.
$L_{i}^{g}$ are expressed in image coordinates normalized so the
face size is constant. 
\item The second set is obtained by further refining $L_{i}^{g}.$ We train
$n$ ANN-Star models $\mathcal{M}_{i}$ (Section \ref{sub:ANN-Star})
, one for each facial landmark. The models are trained for each test
image using its $K$-nearest neighbors. We then use $\mathcal{M}_{i}$
to predict the locations of the facial landmarks. Denote the locations
found by this method as $(L_{i}^{*})_{i=1}^{n}$. 
\end{itemize}
The initial locations $L_{i}^{g}$ are usually within a few pixels
from the correct locations but small inaccuracies are caused due to
different facial proportion, expressions, etc. $L_{i}^{*}$ are more
accurate, except in highly occluded facial regions. Therefore, we
set the location of each landmark as:

\begin{equation}
L_{i}=\begin{cases}
L_{i}^{*} & d_{i}\le T_{L}\\
L_{i}^{g} & d_{i}>T_{L}
\end{cases}
\end{equation}
Where $d_{i}=\||L_{i}^{*}-L_{i}^{g}\|_{2}$ and $T_{L}=30$ (30 pixels,
roughly $1/3$ of the size of the normalized face image) is a threshold
kept constant in all experiments. The refinement increases accuracy
significantly, for quantitative results see Fig. \ref{fig:landmark localization results}
and Section \ref{sec:Experiments}. We have trained our model to detect
seven facial landmarks, which are: the left eye center, right eye
center, left \& right corners of mouth, mouth center, tip of nose
and chin. 

\section{Additional Annotations \& Training}

In this section, we describe our training procedure. As our training
requires additional landmarks, we first describe (Section \ref{sub:Adding-to-the})
how we augment the Standford-40 dataset with additional manual annotations.
Then we describe \ref{sub:Training} how we use these additional annotations
to train the classifier described in Section \ref{sec:Approach}.

\subsection{Adding to the dataset (FRA-db) \label{sub:Adding-to-the}}

Our method requires for training additional information other than
the image label, namely: the bounding box of the face of the person
performing each action, the locations of the 7 facial landmarks (see
\ref{sub:Facial-Landmarks}) and the locations of the action objects,
given as region masks. We augmented the Stanford-40 Actions dataset
\cite{ICCV11_0744} with additional manual annotations. For each image
in the original dataset, we mark the face of the person performing
the action using a bounding box. In rare cases where there are multiple
people performing the action, we mark their faces as well. If the
person's face is either occluded or not visible, we mark the expected
location of the face. For a subset of categories including the 4 face-related
actions, we further annotate the following:

1. 7 facial landmarks: the left eye center, right eye center, left
\& right corners of mouth, mouth center, tip of nose and chin. 

2. A polygonal contour delineating the action-object, except for images
where it is not visible, due to occlusion (e.g by the person's hand,
smoke occluding a cigarette, etc.).

In addition to the 4 classes mentioned above, we also added these
annotations to phoning (which we did not consider in this study).
For the rest of the classes, we also include facial landmarks, which
are extracted automatically as described in Section \ref{sub:Facial-Landmarks}.
As the training procedure (see next section) requires the faces and
facial landmarks, we use the automatically extracted ones as \textquotedblleft ground
truth\textquotedblright{} - except where the face-detector's score
was below 0, at which point its precision drops significantly. In
the future we intend to extend the manual annotations to the entire
dataset (where relevant - i.e, landmarks for visible faces and hands
and objects for transitive actions only). Naturally, we use the annotations
only during training. 

Our augmentation of the dataset also contains some further annotations
not currently used, which are:
\begin{itemize}
\item A \textquotedblleft don't care\textquotedblright{} flag for each facial
landmark, indicating that it invisible due to self occlusion (pose
of face) or occlusion by an object. The occlusion flags are not used
in our algorithm. We used them only for evaluating the accuracy of
our landmark localization (Section \ref{sub:Facial-Landmarks})
\item A bounding box around the hand(s) performing the action
\end{itemize}
The annotations will be released alongside this paper.

\subsection{Training \label{sub:Training}}

We next describe our training procedure. To be robust to inaccuracies
of the segmentation algorithm, and diversify our pool of positive
examples, we started from the set of ground-truth action objects regions
and extended them: for each training image $I_{F},$ let $R$ be the
pool of candidate regions we produce as in Section \ref{sub:Detecting-Action-Objects},
and let $r_{gt}$ be the ground-truth mask of the action-object. We
compute the overlap score of each $r\in R$ with $r_{gt}$, $\ie$,
$Ovp(r,r_{gt})=\frac{|r\cap r_{gt}|}{|r\cup r_{gt}|}$. Now we define
\begin{equation}
R^{+}=r_{gt}\cup\{r:Ovp(r,r_{gt})\ge T_{+}\}
\end{equation}
and 
\begin{equation}
R^{-}=\{r:Ovp(r,r_{gt})\le T_{-}\}
\end{equation}

If $I_{F}$ belongs to the current positive class, we treat $R^{+}$
as positive examples and $R^{-}$ as negatives. If $I_{F}$ belongs
to the negative class, we treat both $R^{+}$ and $R^{-}$ as negatives.
We set $T_{+}=0.55$, $T_{-}=0.3$. We extract the appearance and
interaction features $V_{Obj}$, $V_{Int}$ from all regions collected
in this manner from the training images. The final representation
of each segment is the concatenation of $V_{Obj}$ and $V_{Int}$.

A linear SVM is trained using this representation in a one-vs-all
manner for each class. The score output by the SVM is a linear function
of its parameters, so the scores $\mathcal{\mathcal{\eta}}_{c}^{Int}$
and $\eta_{c}^{Obj}$ (Eq. \ref{eq:energy function}) can be recovered:
let us split the weight vector $w$ learned by the SVM into two parts
corresponding to the features in $V_{Int}$ and $V_{Obj},$ $\ie$,$w=(w_{Int},w_{Obj})$.
Then (assuming the bias is 0 for simplicity), we have $\mathcal{\mathcal{\eta}}_{c}^{Int}=w_{Int}^{T}V_{Int}$
and $\mathcal{\mathcal{\eta}}_{c}^{Obj}=w_{Obj}^{T}V_{Obj}$. The
scores $\eta_{c}^{F}$ and $\eta_{c}^{M}$ in Eq. \ref{eq:energy function}
are trained by extracting fc6 features from the ground-truth image
windows of faces and mouths, respectively (Section \ref{sub:Face-and-Mouth}).
We concatenate these fc6 features from $I_{M}$ and $I_{F}$ into
a single feature vector and train a linear SVM over this representation.
As before, this allows us to express the final output of the classifier
as two scores, relating to $\eta_{c}^{F}$ and $\eta_{c}^{M}$. 

The training stage also includes training an ANN-Star model which
is applied to produce a location map, used as one of the interaction
features (see \ref{sub:Person-Object-Interaction}). It is trained
by recording the center of the action-object in each training image
and producing a model to predict it in test images, using features
from $I_{F}$. As this prediction serves as one of the interaction
features used in the SVM, we need to extract it from the training
images as well. We do so for each training image by voting with all
the training features except those extracted from that image, which
is almost identical to the full model. The ANN-Star model is trained
simultaneously for all considered action classes. The common training
produced better results than training for each action class independently,
possibly due to the relatively low number of training samples. 

The final score is computed using Eq. \ref{eq:energy function}, summing
the outputs of the different classifiers for the face, mouth and action
object. The constant $\gamma$ was set to $0.1$ to accommodate for
the larger dynamic range of $\mathcal{\mathcal{\eta}}_{c}^{Int}+\eta_{c}^{Obj},$
which was approx. 10 times as large as that of $\eta_{c}^{F}$ +$\eta_{c}^{M}$. 

The entire classification process is summarized in Alg. \ref{alg:Classifying-Face-Related-Actions}.
\begin{algorithm}
\protect\caption{\label{alg:Classifying-Face-Related-Actions}Classifying Face-Related
Actions}

\begin{enumerate}
\item Given a test image $I$ and target class $c$, apply face detection
inside bounding box of person
\item $X,S\leftarrow$bounding box and score of best scoring face in I
\item If $S<S_{min}$ return $-inf$
\item $I_{F}\leftarrow$crop $I$ with $X$ inflated by 1.5 
\item $L$ = $DetectLandmarks(I_{F})$ (Section \ref{sub:Facial-Landmarks});
$F\leftarrow(X,S,L)$
\item Extract fc6 features from $I_{F}$ and $I_{m}$ (Section \ref{sub:Face-and-Mouth})
and use them to calculate $\eta_{c}^{F}(I_{F})$ and $\eta_{c}^{M}(I_{M})$ 
\item Create a pool of candidate action-object regions $R$ (Section \ref{sub:Detecting-Action-Objects})
\item For each $r\in R$

\begin{enumerate}
\item compute interaction features $V_{Int}$ and object features $V_{obj}$
\item compute $\mathcal{\mathcal{\eta}}_{c}^{Int}(r,F)$ and $\eta_{c}^{Obj}(r,I_{F})$
using learned classifier
\end{enumerate}
\item $s_{c}(R)\leftarrow\max_{r\in\mathcal{R}}(\mathcal{\mathcal{\eta}}_{c}^{Int}(r,F)+\eta_{c}^{Obj}(r))$
\item Return $S_{c}(I,F)=\eta_{c}^{F}(I_{F})+\eta_{c}^{M}(I_{M})+\gamma s_{c}(R)$ \end{enumerate}
\end{algorithm}


\section{Experiments\label{sec:Experiments}}

\begin{figure*}
\protect\caption{\label{fig:Interp-image}Identifying the action object. In addition
to classifying the action being performed, we output a delineation
of the image region (outlined in red) which scored best out of the
pool of candidates considered. Last two columns: failure cases, where
the action object is either found inaccurately (top-right) or missed
due to a wrong candidate having a higher score (bottom-right; note
the small real cigarette stud between the tip of the index and middle
fingers). Best viewed online.}

\begin{centering}
\includegraphics[width=0.15\textwidth]{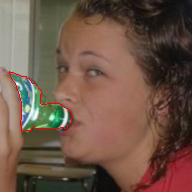}\includegraphics[width=0.15\textwidth]{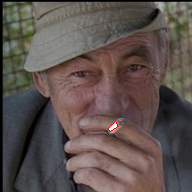}\includegraphics[width=0.15\textwidth]{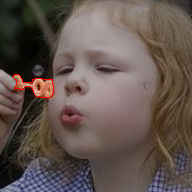}\includegraphics[width=0.15\textwidth]{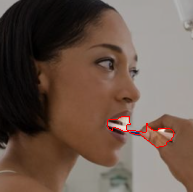}\includegraphics[width=0.15\textwidth]{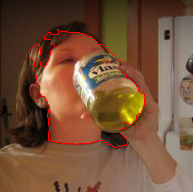}
\par\end{centering}

\centering{}\includegraphics[width=0.15\textwidth]{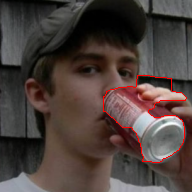}\includegraphics[width=0.15\textwidth]{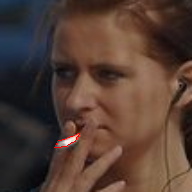}\includegraphics[width=0.15\textwidth]{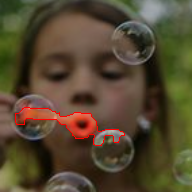}\includegraphics[width=0.15\textwidth]{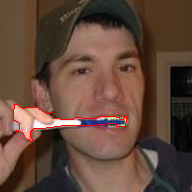}\includegraphics[width=0.15\textwidth]{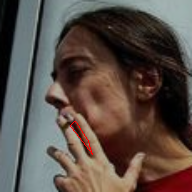}
\end{figure*}

In the following we describe experiments evaluating the algorithm,
including measuring performance of our method on the Stanford-40 dataset
in different settings, with some visualization of its output, as well
as testing our facial-landmark detection.

\subsection{Facial Landmarks}

To examine the effectiveness of our facial landmark detection, we
tested it on the part of FRA-db annotated with facial landmarks. These
include overall 1215 images. The localization method was currently
trained to detect 7 facial landmarks (Section \ref{sub:Facial-Landmarks}).
We compare with the method of Zhu et al \cite{zhu2012face}, as their
method produces landmarks for profile faces as well, unlike many others
who only consider near-frontal faces. Using landmarks common to both
methods (which includes all our landmarks except the center of the
mouth). In order to remove cases where the face is not detected at
all, we provide the bounding box of the face for both methods. There
are still some cases where their detector produces no result (this
happens when their method arrives at a final score lower than a set
threshold, -0.9), which we discard for our comparison. We also ignored
landmarks which were marked in our annotation (Section \ref{sub:Adding-to-the})
as ``don't care'' (due to being invisible due to self occlusion,
etc) for evaluation purposes. Normalizing by the size of the face,
we computed the distribution of average errors for each method. Results
can be seen in Fig. \ref{fig:landmark localization results}. Overall
our method is more robust to large occlusions, which is important
for finding accurately the configuration of objects w.r.t facial landmarks.

\subsection{Action Recognition}

We tested on the Stanford 40-actions dataset \cite{ICCV11_0744},
using the same train-test split. It contains 4000 train and 5532 test
images, 100 training images for each of the categories. For each image
we are also given the bounding box of the person performing the action.
For interaction features, which required the location of the face
and facial landmarks, we used the annotations in FRA-db at train time.
At test time we used the face detection of \cite{mathias2014face}
inside the bounding box containing the person. In each test image,
we took as a face the single best face detection, where an image whose
best scoring face was below $0$ was classified as non-class. Given
the face detections we extracted facial landmarks as in Section \ref{sub:Facial-Landmarks},
which enabled us to compute the rest of the features, as described
in Sections \ref{sub:Detecting-Action-Objects} and \ref{sub:Person-Object-Interaction}.
We applied the trained classifier to each putative face, and then
computed the average precision over the entire dataset. Table \ref{tab:Performance-comparision-of}
gives summary of the results as well as a comparison with recent state-of-the-art
results. 

To simulate an ideal face/head detector, we ran our classification
again, this time using our manual annotations of \textit{all} faces
in the test set. Many of these include faces that are either small,
facing away from the camera, and other challenging settings. The results
are given in the last column of Table \ref{tab:Performance-comparision-of}.
To encourage others to test their performance on this diverse and
challenging set of faces, we make it publicly available.

\begin{table}
\begin{centering}
\begin{tabular}{|c|c|c|c|c|c|}
\hline 
 & \cite{ICCV11_0744} & \cite{khan2013coloring} & \cite{shahbaz2013semantic} & Ours & Ours{*}\tabularnewline
\hline 
\hline 
Drinking & 20 & 13 & 19 & \textbf{43.9} & 44.8\tabularnewline
\hline 
Smoking & 30 & 30 & 39 & \textbf{44.6} & 44.6\tabularnewline
\hline 
Blowing Bubbles & 40 & 43.5 & 43 & \textbf{67.0} & 68.3\tabularnewline
\hline 
Brushing Teeth & 39 & 32 & 36 & \textbf{52.6} & 57.4\tabularnewline
\hline 
Mean & 32.25 & 29.6 & 34.2 & \textbf{52} & 53.8\tabularnewline
\hline 
\end{tabular}
\par\end{centering}

\protect\caption{\label{tab:Performance-comparision-of}Performance comparison of avg.
precision of the proposed method vs. recent state-of-the-art results
on 4 face-related actions in the Stanford-40 Actions dataset .\textbf{
Ours} is for the results of the proposed method.\textbf{ Ours{*} }is
with face locations provided by manual annotation.}
\end{table}

\subsection{Interpretability of Results}

A desirable property in a recognition algorithm is that in addition
to being able to correctly label images, it will be able to explain
\textit{why }it has done so. Our algorithm seeks the action object
which is valid both in configuration relative to the face and in its
appearance. An outcome of the current approach is that it produces
an estimation of the action objects, which is a partial explanation
of why it has reached its decision for the image being considered.
Some examples of the best scoring regions (out of thousands of candidates)
can be seen in Fig. \ref{fig:Interp-image}.

\section{Conclusions and Future Work}

We have presented a method for analyzing transitive actions, specifically
related to the face. Our method is based on identifying the relevant
body parts (\eg, face, mouth) and the action-object, and extracting
features from the body part, action-object, and their interaction.
It can be generalized to other actions by using models based on different
body-part and action-object pairs. The method improves over existing
methods by a large margin, as well as providing interpretable results.
As mentioned in the introduction and illustrated in Fig. \ref{fig:action-zoomed-in},
disambiguation between similar actions depends on fine specific shape
and appearance details at specific locations. Therefore, we represented
the shape of each region in several ways, to capture fine features
that proved as informative for disambiguation. The set of useful shape
and interaction features can be further extended in future studies.
One option is to start with a large pool of candidate feature types
and retain those which are found to be most informative. In the future,
we intend to expand our analysis to additional body parts and actions,
and deal with more subtle details regarding the appearance of the
object itself. Finally, we make public our augmentation of an existing
dataset for general use.

\bibliographystyle{plain}
\bibliography{arxiv_2015}

\end{document}